\documentclass[preprint,12pt]{elsarticle}

\usepackage{amssymb}
\usepackage{float}
\usepackage{hyperref}

\journal{CoRR}

\begin{document}

%    \makeatletter
%    \def\ps@pprintTitle{%
%       \let\@oddhead\@empty
%       \let\@evenhead\@empty
%       \let\@oddfoot\@empty
%       \let\@evenfoot\@oddfoot
%    }
%    \makeatother

\begin{frontmatter}



\title{Using accumulation to optimize deep residual neural nets}


\author{Yatin Saraiya}

\address{847 Moana Court, Palo Alto 94306, CA, USA}
\ead{yatinsaraiya12@gmail.com}

\begin{abstract}
Residual Neural Networks \cite{residual}  won first place in  
all five main tracks of the ImageNet and COCO 2015 competitions.
This kind of network involves the creation of pluggable modules such that the output contains a residual from
the input.  The residual in that paper is the identity function.  We propose to include residuals from all
lower layers,
suitably normalized, to create the residual.  This way, all previous layers contribute equally to the
output of a layer.  We show that our approach is an improvement on \cite{residual} for the CIFAR-10 dataset.

\end{abstract}


\end{frontmatter}


\section{Introduction}
\label{intro}
Deep convolutional neural networks \cite{CNNs,backpropagation} form the basis for image recognition.  
It has been shown \cite{Simonyan} that depth is critical in classification accuracy.  The stacked
layers of such nets provide features at different granularities \cite{features}.   However, very deep
neural nets suffer from degradation of training error as the networks start converging.
Proposed solutions to this degradation problem include shortcutting \cite{bishop}, of which the use of residuals as in 
\cite{residual} is a modification.
These networks consisted of stacked blocks with the same input-output characteristics\footnote{modulo a small
number of changes in the layer's input and output sizes.}, with
residuals from the input added to the output via the identity function.

\begin{figure*}
  \includegraphics[width=0.2\textwidth]{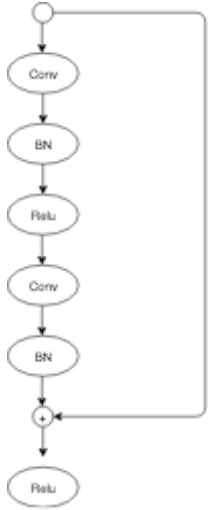}
\caption{Residual block}
\label{residual}       % Give a unique label
\end{figure*}
\begin{figure*}
  \includegraphics[width=0.2\textwidth]{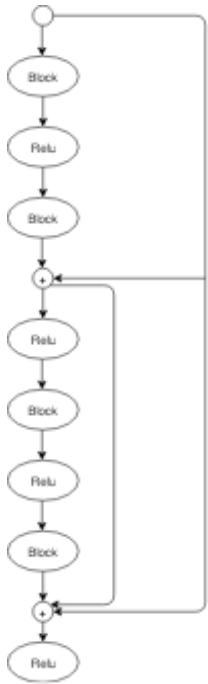}
\caption{Accumulated residual block}
\label{residual_new}       % Give a unique label
\end{figure*}
Figure~\ref{residual} illustrates one such block.  The results of \cite{residual} showed
that this modular design would mitigate degradation even in very deep networks.
The fundamental improvement was the addition of residuals using the identity function.
That is, if $F$ is the function computed by a block, $x$ is the input and $y$ is the output,
\begin{equation}
\label{identity}
y = \sigma(F(x) + x)
\end{equation}
where $\sigma$ is the rectified linear unit.
The intuition is that later layers perform fine tuning on the results of the earlier layers.

We replace the identity residual with the sum of the normalizations of the inputs to each
block, which necessitates just one extra variable that accumulates the residue, and one extra
addition per block\footnote{We reinitialize whenever the input to a block changes shape, which
removes the necessity for the addition.}. 
If the model consists of blocks $B_1,B_2,\ldots,B_L$, $F_i$ is the function computed by block $B_i$,
$x_i$ is the input to block $B_i$
and $y_i$ is the output of this block, we have
\begin{equation}
\label{normalized}
y_i= \sigma(F_i(x_i) + \Sigma_{j=1}^{i}\textnormal{BN}(x_i))
\end{equation}
where $\textnormal{BN}(x_i)$ is the batch normalization of $x_i$.  The intuition is that each block computes feature sets
at a different granularity, so each block's output should weigh equally in the result.
 Figure~\ref{residual_new} presents the architecture. We call such neural nets \begin{it}{accumulated residual neural nets}\end{it}.

\section{Experiments}
\label{experiments}
We used \texttt{cifar10\_resnet.py}, obtained from 
\newline\texttt{https://github.com/fchollet/keras/blob/master/examples/},
which bears the MIT license, as a representation of the network described in \cite{residual}.
We modified it to define the network of this paper.  

We ran both against the CIFAR-10 dataset \cite{cifar10} with the depth at 32.  The experimental setup was
that of Section~4.2 of \cite{residual}.
Note that the same setup was used for both the residual network and our network.

\paragraph{Results}
%The time per epoch for the residual net with history was 15\% higher than the time per epoch of the residual net.

\begin{figure*}
  \includegraphics[width=0.9\textwidth]{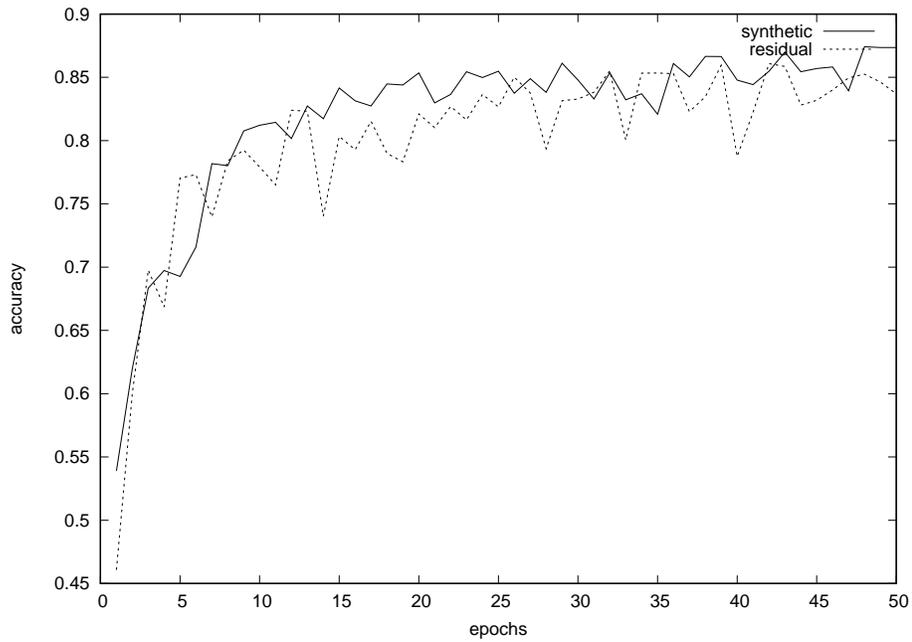}
\caption{Validation accuracy}
\label{residual4}       % Give a unique label
\end{figure*}

\begin{figure*}
  \includegraphics[width=0.9\textwidth]{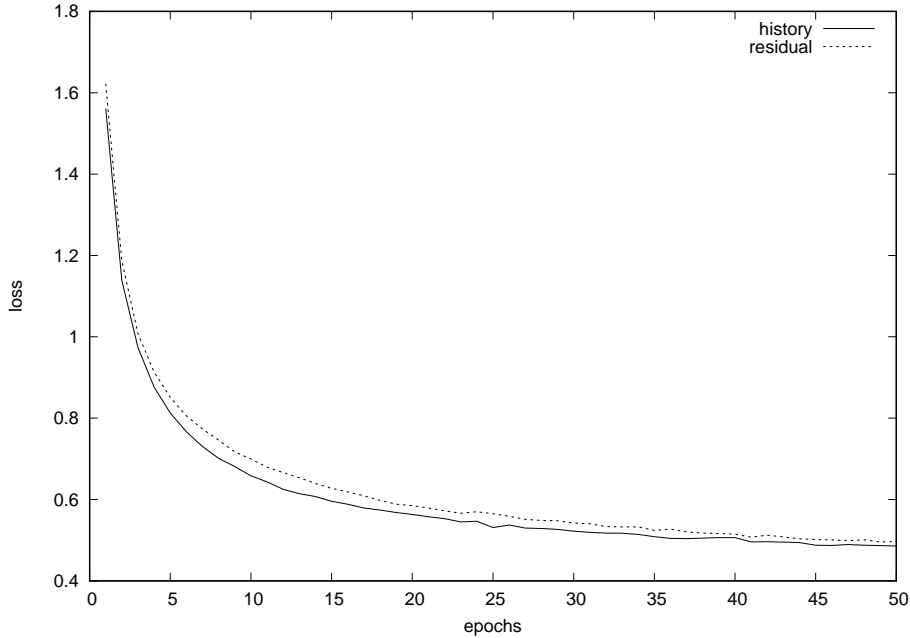}
\caption{Training loss}
\label{residual1}       % Give a unique label
\end{figure*}

\begin{figure*}
  \includegraphics[width=0.9\textwidth]{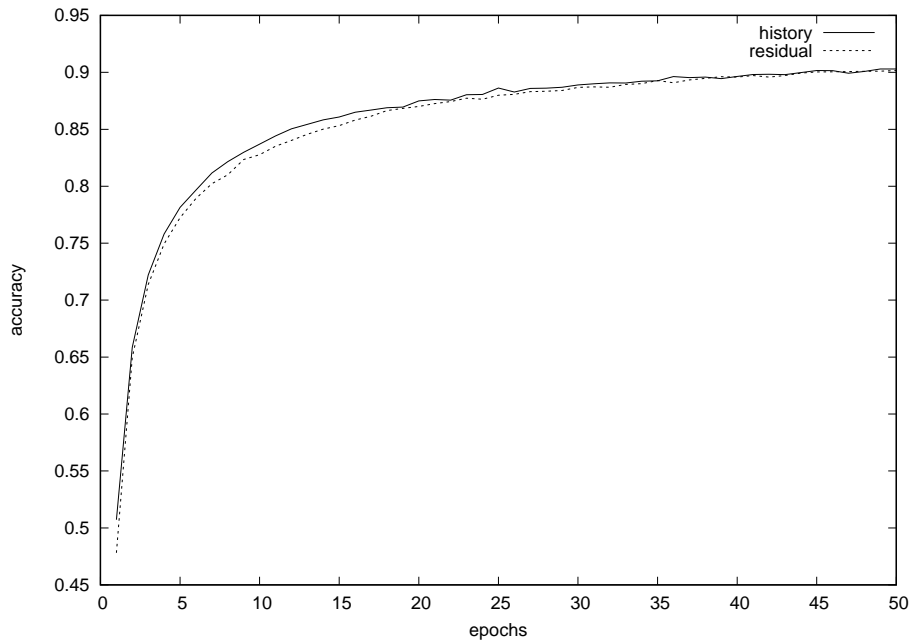}
\caption{Training accuracy}
\label{residual2}       % Give a unique label
\end{figure*}

\begin{figure*}
  \includegraphics[width=0.9\textwidth]{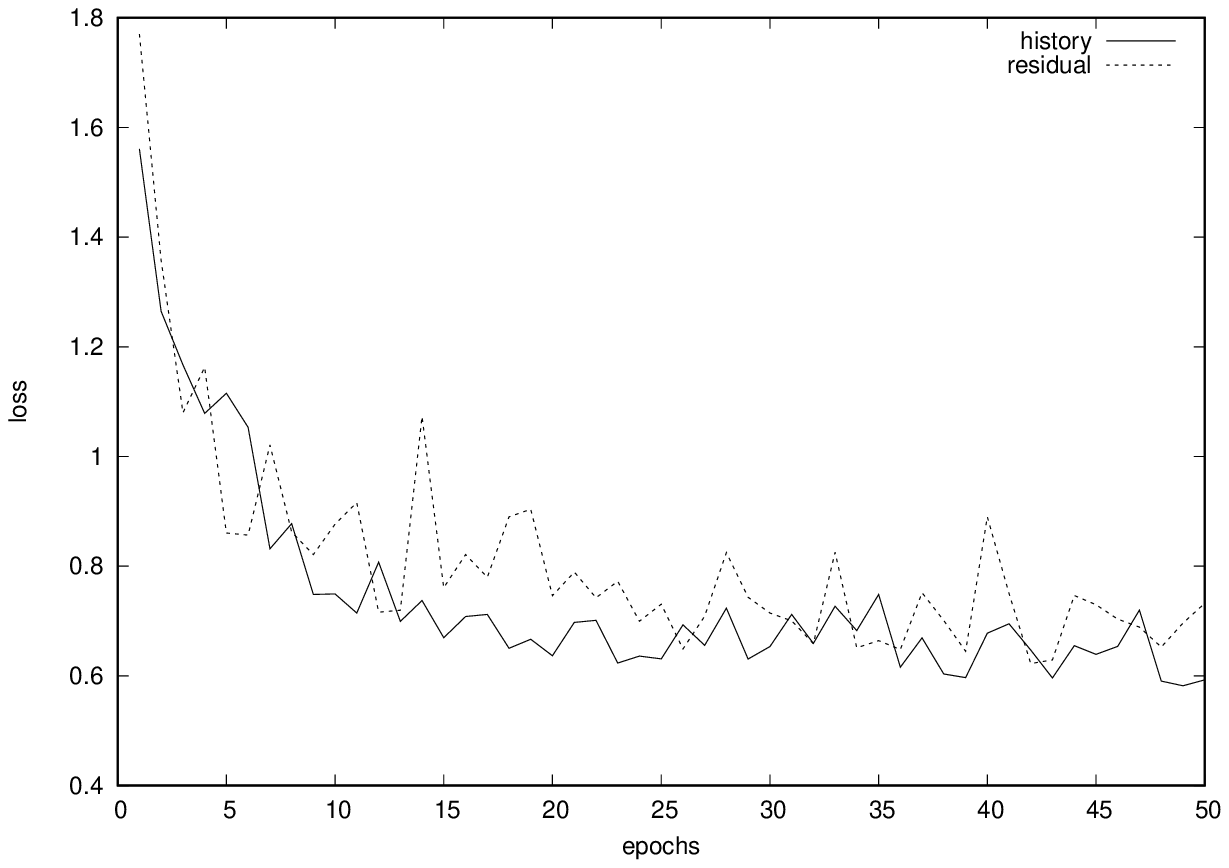}
\caption{Validation loss}
\label{residual3}       % Give a unique label
\end{figure*}

Our results are contained in Table~\ref{resultstable} and
Figures~\ref{residual1}, \ref{residual2}, \ref{residual3} and \ref{residual4}.

\begin{table}
\centering
\label{resultstable}       % Give a unique label
\begin{tabular}{|l|c|c|}
\hline\noalign{\smallskip}
&Min top-1 error& Avg top-1 error\\
\noalign{\smallskip}\hline\noalign{\smallskip}
ResNet & 13.92 & 19.9 \\
Accumulated & 12.65 & 18.1 \\
\noalign{\smallskip}\hline
\end{tabular}
\caption{Top-1 validation error over 50 epochs}
\end{table}

Table~\ref{resultstable} presents the minimum and average validation errors per epoch over 50 epochs.
In each case, the net with history was at leasts 1\% better than the residual net.  That is, it generalizes
better on the CIFAR-10 dataset.

Figure~\ref{residual4} compares the validation accuracy of the residual net with and without accumulation.

Figure~\ref{residual1} compares the training loss of the residual net with and without accumulation.
Figure~\ref{residual2} compares the training accuracy of the residual net with and without accumulation.
Figure~\ref{residual3} compares the validation loss of the residual net with and without accumulation.

\section{Conclusions}
\label{conclusions}
We presented an augmentation of residual neural networks where the residuals accumulate along the depth
of the neural net.  This permits the output of each layer to play an equal role in the classification.
We showed that these networks outperform the residual networks of \cite{residual}.  It is of interest
to see whether this approach extends to the wide networks of \cite{wide} and the aggregated networks
of \cite{aggregate}.





\end{document}